# PSA-Net: Deep Learning based Physician Style-Aware Segmentation Network for Post-Operative Prostate Cancer Clinical Target Volume

Anjali Balagopal[1,2, a], Howard Morgan[1,2,a], Michael Dohopoloski[1,2], Ramsey Timmerman[2], Jie Shan[3], Daniel F. Heitjan[4,5], Wei Liu[3], Dan Nguyen[1,2], Raquibul Hannan[2], Aurelie Garant[2], Neil Desai[2], Steve Jiang[1,2, b]

[1]Medical Artificial Intelligence and Automation (MAIA) Laboratory, University of Texas Southwestern Medical Center, Dallas, Texas, USA
[2]Department of Radiation Oncology, University of Texas Southwestern Medical Center, Dallas, Texas, USA
[3]Department of Radiation Oncology, Mayo Clinic, Phoenix, Arizona, USA
[4]Department of Statistical Science, Southern Methodist University, Dallas, Texas, USA
[5]Department of Population & Data Sciences, University of Texas Southwestern Medical Center, Dallas, Texas, USA
[a]Co-first authors.
[b]Correspondence author. E-mail: Steve.Jiang@UTSouthwestern.edu

# ABSTRACT

**Purpose:** Automatic segmentation of medical images with deep learning (DL) algorithms has proven to be highly successful in recent times. With most of these automation networks, inter-observer variation is an acknowledged problem, leading to sub-optimal results. This problem is even more significant in post-operative clinical target volume (CTV) segmentation due to the absence of macroscopic visual tumor in the image. This study, using post-operative prostate CTV segmentation as the test bed, tries to determine: 1) if physician styles are consistent and learnable; 2) if there is an impact of physician styles on treatment outcome and toxicity; and 3) how to explicitly deal with physician styles in DL-assisted CTV segmentation to facilitate its clinical acceptance.

**Method:** A dataset of 373 postoperative prostate cancer patients from UT Southwestern Medical Center are used for this study. Another 83 patients from Mayo Clinic are used to validate the developed model and its adaptation ability. A 3D convolutional neural network classifier was trained to identify which physician had contoured the CTV from just the contour and corresponding CT scan, to determine if physician styles are consistent and learnable. Next, we evaluated if adapting automatic segmentation to specific physician styles would be clinically feasible based on a lack of difference between outcomes. Here, biochemical progression-free survival and grade 3+ genitourinary and gastrointestinal toxicity were estimated with the Kaplan-Meier method and compared between physician styles with the log rank test and subsequently with a multivariate Cox regression. When no statistically significant difference in outcome or toxicity between contouring styles was shown, we proposed a concept called physician style-aware (PSA) segmentation, by developing an encoder-multidecoder network with perceptual loss to model different physician styles of CTV segmentation.

**Results:** The classification network captured the different physician styles with 87% accuracy. On subsequent outcome analysis, there were no differences between BCFS and grade 3+ toxicity among physicians. With the proposed physician style-aware network (PSA-Net), Dice similarity coefficient (DSC) accuracy increases on an average of 3.4% for all physicians from a general model that does not differentiate physician styles. We show that these stylistic contouring variations also exist between institutions that follow the same segmentation guidelines and show the effectiveness of the proposed method in adapting to new institutional styles. We observed an accuracy improvement of 5% in terms of DSC when adapting to the style of a separate institution.

**Conclusion:** The performance of the classification network established the existence of learnable physician styles, which appear clinically feasible to adapt to, based on a lack of difference between outcomes among physicians. Thus, a novel PSA-Net model was developed, which is capable of producing contours specific to the treating physician, improving segmentation accuracy, and avoiding the need of training multiple models to achieve different style segmentations. This model was further validated on data from a separate institution successfully, supporting its generalizability to diverse dataset.

# 1. INTRODUCTION

Optimal radiation treatment entails uniform full-dose coverage of the radiation target with a sharp dose fall-off around it. This necessitates precise segmentation of both the radiation target and the nearby organs that are at risk for radiation damage (Organs-at-risk, OARs). In a typical radiotherapy setting, OARs are segmented together with the gross tumor volume (GTV), which is the tumor that is visible in images. Using their knowledge of the disease, physicians then expand the GTV to create the clinical target volume (CTV), which also includes the microscopic extensions not visible in images. In the case of post-operative radiotherapy, however, the visible tumor has been surgically removed, so the CTV is only a virtual volume encompassing area that may contain microscopic tumor cells, not an expansion of a macroscopic or visible tumor volume. CTV definition in the postoperative setting gets even more complicated for many reasons including changes in anatomy caused by the surgery itself and the limited information on the preoperative location of the prostate. Consequently, segmenting the CTV in these cases, whether manually or automatically, is much more challenging than segmenting typical organs or CTVs expanded from GTVs.

Given the absence of a visible structure, physicians follow consensus guidelines for contouring the CTV. Five published consensus guidelines have variously defined the post-operative CTV in prostate cancer radiotherapy (European Organization for Research and Treatment of Cancer [EORTC][1], Faculty of Radiation Oncology Genito-Urinary Group [FROGG][2], Princess Margaret Hospital [PMH][3], Radiation Therapy Oncology Group [RTOG][4] and Francophone Group of Urological Radiotherapy [GFRU][5]) ; however, there is no universally accepted method of segmenting the CTV. It is also recognized that physicians must consider not only published guidelines, but individual patient characteristics, including the presence of seminal vesicle invasion, positive margins, variations in anatomy, and co-morbidities, when segmenting the CTV. There is a lot of room for physicians to exercise their clinical judgment for individual patients, based on their training backgrounds, experiences, and personal preferences, which can lead to large variations among physicians[6-11] within an institution as well as across institutions.

In most automated organ/tumor segmentation studies, observer variation is an acknowledged problem as it makes training data heterogeneous. This problem becomes more significant in post-operative cases where there is an absence of visible ground truth and treating physician has an upper hand in deciding which contour to treat as long as guidelines are followed. Significant variability exists between radiation oncologists with respect to prostate bed CTV delineation in postoperative patients[6-11]. Specifically, Symon and colleagues[10] found that the interobserver variability for postoperative CTV delineation between five radiation oncologists on a sample of eight patients varied from 39 to 53 $cm^3$ for the patient with the smallest and from 16 to 69 $cm^3$ for the patient with the largest variation. These studies showed that, even though published guidelines are available to standardize postoperative prostate CTV delineation, significant variations between physician contours still exist.

In a previous study[12], our group developed a deep learning (DL) based auto-segmentation tool that effectively auto-segmented CTV. We observed that physicians slightly preferred their own contours to AI contours and clearly preferred AI contours over other physicians' contours. It was also observed that the majority of the edits required for the contours that did not receive a perfect score were stylistic in nature. If the AI model can mimic a physician's style, then the accuracy of the DL models can be further improved. Since it is difficult, if not impossible, to bring a consensus for all physicians in the foreseeable future, individual physician segmentation styles have to be respected and modeled when developing automatic segmentation tools. Doing so can not only increase automated contour acceptance but can also display the possible contouring variations to the attending physicians, potentially reducing variability between physicians. Reduced variation in the definition of the CTVs can make for a more robust evaluation of new radiation therapy techniques particularly when multiple institutions are involved.

Inter-observer variability in automated segmentation is mentioned as a limitation in many medical image segmentation-related papers. One proposed solution is to have multiple observers contour on every patient and agree on a contour for training. This increases segmentation accuracy but does not explicitly solve the problem. When deployed in the clinic, physicians would have to spend time correcting these auto contours for stylistic changes, in addition to verifying the margins of the CTV in relation to surrounding OARs and inclusion of at-risk structures based on patient factors. Moreover, such data is rarely available. Another method by which a highly accurate model is trained, is by using only single observer data for training. This significantly reduces the amount of data that can be used for training and produces segmentations in a single physician's style. However, the scope of application of such a trained model is limited. So far, no attempts have been made to explicitly capture observer variation and to train a deep learning model that can mimic multiple desired styles.

In this work, we study the inter-physician CTV contouring style variations using deep learning. We first train a convolutional neural network (CNN) to capture the different style variation of physicians, which we then use to identify which physician has segmented a particular CTV. We also conduct an outcome analysis study to analyze the impact that physician styles have on patient outcome. This study investigates the effect of physician style on biochemical recurrence-free survival and treatment toxicity. If a significant clinical impact can be established, the contours of physicians who have better outcomes can be used to improve published CTV contouring guidelines in the future.

In situations where there is no significant impact on treatment outcome, we propose a new concept called physician-style aware (PSA) segmentation, by developing a new architecture, PSA-Net, an encoder-multidecoder network capable of mimicking different styles of segmentation. PSA-net can potentially improve quality and clinical acceptance of automatic segmentation methods. The proposed method additionally uses a novel perceptual loss function that can capture stylistic variations between physicians, thus creating visually different physician CTV contours. We also show that these stylistic variations exist between institutions that follow the same segmentation guidelines, and then show the effectiveness of the proposed method in adapting to new institutional styles.

## 2. METHODS

## 2.1 DATA

### 2.1.1. SEGMENTATION

**UTSW Dataset**

Our dataset includes retrospectively collected CT volumes obtained at initial simulation for adjuvant or salvage postoperative radiotherapy (PORT) to the prostatic fossa of 373 patients who initially underwent resection for prostate cancer at UT Southwestern Medical Center (UTSW). These scans were contoured by 4 different physicians, and these contours, henceforth called *clinical contours*, were used for the treatment of these patients. At the UTSW genitourinary (GU) radiation oncology service, these clinical contours are usually drawn by residents, corrected by supervising attending physicians, and reviewed by all attending physicians. All physicians used the RTOG guideline as the reference[4]. Each CT volume contains 60-360 slices and a voxel size of $1.17 \times 1.17 \times 2$ mm$^3$. We present distributions of volume and stage across physicians in Figure 1. We set aside 60 patients, 15 from each physician, for testing; we used the rest for training and validation. A violin plot showing CTV volume variations across physicians appears in Figure 1.

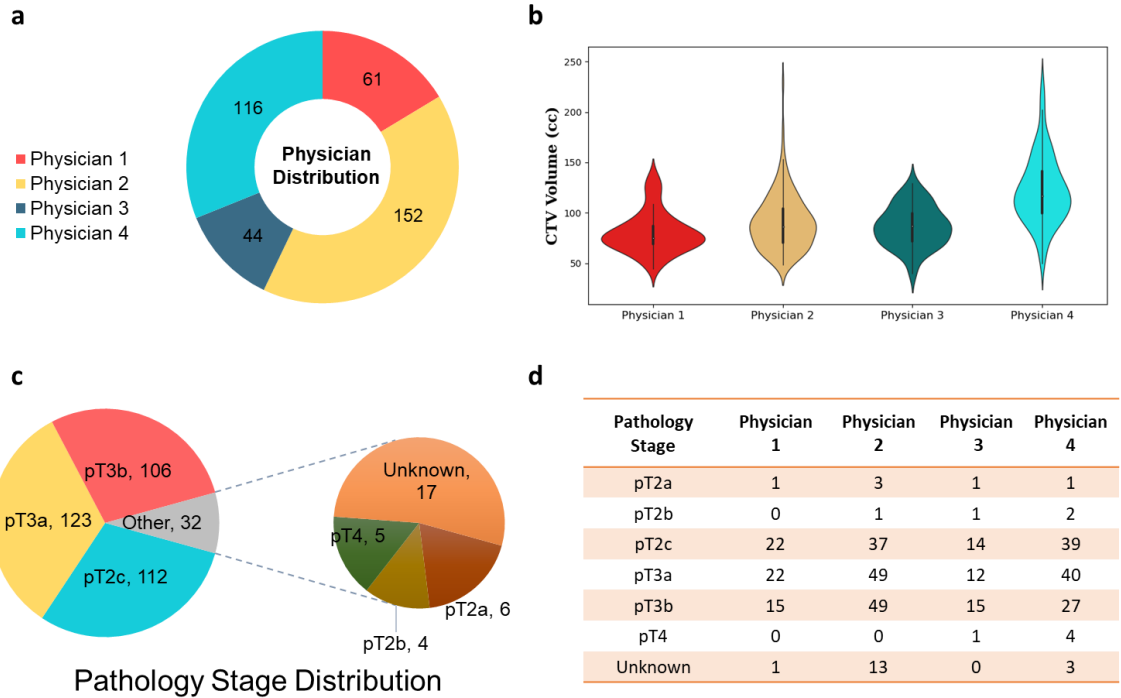

*Figure 1: 373 post-operative prostate cancer patients were used in this study. a) The distribution of patient across 4 different physicians included in this study b) The violin plot showing the distribution of CTV volumes segmented by each of the 4 physicians. c, d) The pathology stage distribution for all the patients*

**Mayo Dataset**

A dataset consisting of 83 patients with the same inclusion criteria, collected at Mayo Clinic in Arizona, was used as an external dataset for evaluation. The initial clinical contours were contoured by 6 different GU radiation oncologists, of which 67 were contoured by a single physician (Figure 2b). RTOG guidelines were used as the reference for CTV segmentation[4]. The CTV volume distributions of the UTSW and Mayo datasets appear in the violin plot in Figure 2a. Mayo MDs make stylistic edits to the RTOG guidelines along the pubic symphysis, which results in smaller CTV volumes.

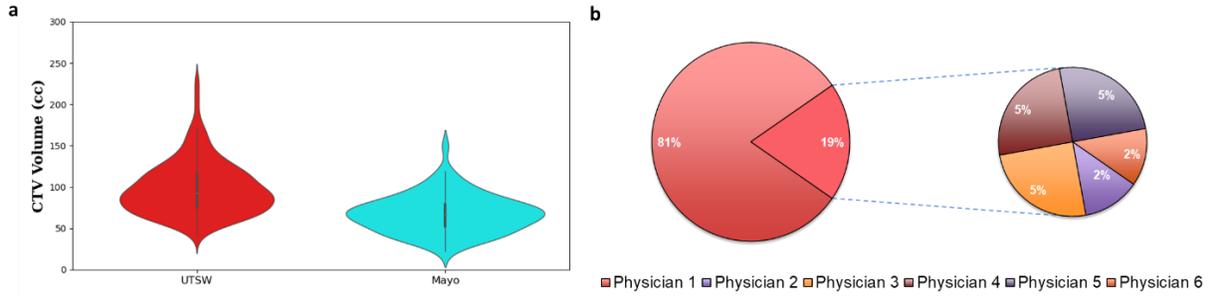

*Figure 2: a) Violin plot showing the volume difference between post-operative prostate CTVs treated in UTSW when compared to Mayo b) Distribution of patients among the 6 different physicians in Mayo dataset.*

### 2.1.2. OUTCOME ANALYSIS

The outcome analysis study included patients treated with PORT by four expert GU radiation oncologists at UTSW from January 2010 to December 2017. All participants were obtained from the UTSW dataset, with additional inclusion criteria being that PORT was delivered only to the prostate bed (n =212). We excluded patients who had radiation delivered to regional lymph nodes (n=166) in addition to prostate bed or who did not have available outcome/toxicity data (n=13). Median follow-up for the included cohort was 4.3 years (interquartile range: 2.5 years to 6.4 years).

### 2.2 CLASSIFICATION NETWORK

To validate the presence of different contouring styles among physicians at UTSW and to evaluate the capability of a neural network to differentiate between styles, we use a 3D deep neural network as a classifier. The goal of this classifier is to classify a CTV contour to a specific physician. A distance map of the CTV contour is given as input to the classifier along with the CT image, and the model predicts the probability of the contour belonging to a specific physician. The distance map is a Euclidean distance transform of the CTV binary mask. The classification network is a CNN network with inception blocks at every layer (Figure 3) . The final layer with SoftMax activation gives the classification output.

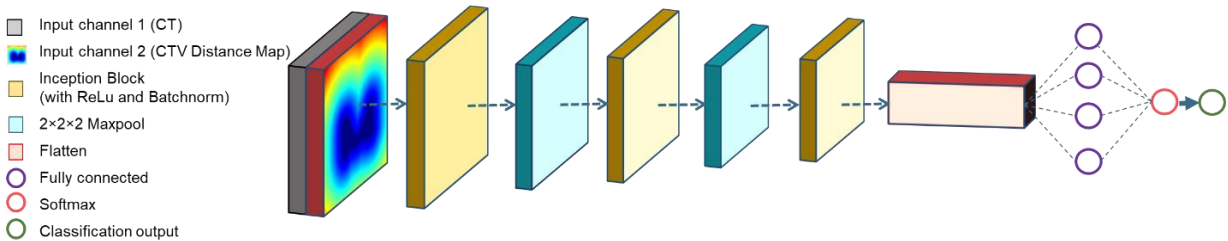

*Figure 3: Classification network for physician style classification. The network takes patient CT and CTV distance map as inputs and predicts the contouring physician.*

### 2.3 STYLE ADAPTATION ARCHITECTURE

Creating style-aware adaptation by training a separate model for each physician, although intuitively appealing, has two main disadvantages. (1) not all patient data are seen by models, reducing data diversity

;(2) some physicians might not have sufficient data for training. Thus, it is preferable to train a model with large datasets using different physician contours as input yet being able to produce contours specific to that physician utilizing this model. For this, we propose PSA-Net with an encoder-multidecoder architecture with feature attention and perceptual loss. Figure 4 depicts the model. As the name suggests, the model has a single encoder and multiple decoders. Each of the decoders is responsible for producing a specific style-aware segmentation. During training, the encoder and the mixed style decoder weights are updated for all the data. For the rest of the physician style decoders as well as the feature attention layers belonging to each decoder, the weights are updated only when the data belongs to a specific physician style. With the help of feature attention, only specific encoded features at every layer that are relevant for a specific physician style are concatenated to the corresponding decoder layer. Having a mixed-style decoder that is updated for all data stabilizes the training; it can also provide a general segmentation if a specific style is not required.

For physician-style decoders, we applied the perceptual loss in the global loss function. High-level perceptual losses[13] that involve a fixed pre-trained network, such as ImageNet pre-trained VGG-19[14] network have been found to be very useful in the frame of style transfer and image synthesis. As opposed to directly measuring the difference between two images, this method calculates the semantic difference between two images as the difference of their feature representations as computed by the fixed network. However, this method cannot be directly translated to medical image segmentation, as natural images and medical images have different numbers of channels and belong to very different distributions. Moreover, because the goal is to generate style-adapted CTV binary segmentation masks, it would be beneficial if visual features that are expected to be different between physicians could be used for calculating perceptual loss. For CTV, visual features that vary between physicians are the shape and the extent of overlap with organs. To ensure that these visual features are preserved in physician style-aware CTV segmentation, we propose the use of a pre-trained ShapeNet and OverLapNet to calculate the perceptual loss during training inspired from semantic edge-aware perceptual loss proposed by Chen et al.[15]. ShapeNet is pre-trained to predict semantic edges from the ground truth CTV segmentation masks and OverLapNet is pre-trained to predict CTV overlap areas with organs from the ground truth CTV and organ segmentation maps. The features of ShapeNet and OverLapNet thus encode information of segmentation masks that correspond to the CTV shape and organ-overlap respectively, and then are used to define the perceptual loss function.

For creating semantic edge maps, given CT and its corresponding ground truth segmentation CTV mask G, we generate a binary ground truth semantic edge map $G_{edge}$ by setting all pixels that do not have 8 identically labeled neighbor pixels as 1, and other pixels as 0. These ground truth semantic edge maps are calculated beforehand and no further computation is needed afterward. Similarly, for creating organ overlap maps, given CT, corresponding bladder and rectum binary masks and the ground truth CTV segmentation mask G, we generate a binary organ-overlap map $G_{overlap}$ by multiplying the CTV mask with the organ masks. These ground truth organ-overlap maps are also calculated beforehand and no further computation is needed afterwards. The pre-trained ShapeNet ($S_\phi$) as well as OverLapNet ($O_\phi$) layers are frozen during training of physician style-aware network. For a patient datapoint, $p0$, the perceptual loss is defined as

$$L_{perceptual,S_\phi}(GT^{p0}, P^{p0}) = \lambda_s (L_{DSC}\{S_{\phi L}(GT^{p0}), S_{\phi L}(P^{p0})\} + \sum_{l=1}^{L-1} \|S_{\phi l}(GT^{p0}) - S_{\phi l}(P^{p0})\|_2 \quad (1)$$

$$L_{perceptual,O_\phi}(GT^{p0}, P^{p0}) = \lambda_O (L_{DSC}\{O_{\phi L}(GT^{p0}), O_{\phi L}(P_i)\} + \sum_{l=1}^{L-1} \|O_{\phi l}(GT^{p0}) - O_{\phi l}(P^{p0})\|_2 \quad (2)$$

Where $L$ is the total number of layers in ShapeNet and OverLapNet and $\lambda_s, \lambda_o$ is a hyperparameter representing the importance of each network on the total loss.

With perceptual loss included, the total loss for training the model will be,

$$L_{total} = L_{Decoder1}^{DSC}\{GT, P_1\} + L_{Decoder,2....N} \tag{3}$$

$$L_{Decoder,2....N} = L_{Decoder,2....N}^{DSC} + L_{Decoder,2....N}^{percept} \tag{4}$$

$$L_{Decoder,2....N}^{DSC} = \sum_{n=1}^{N} \lambda_n^{p0}\left(L_{Decoder1}^{DSC}\{GT^{p0}, P_n^{p0}\}\right) \tag{5}$$

$$L_{Decoder,2....N}^{percept} = \sum_{n=1}^{N} \lambda_n^{p0}\left(L_{perceptual,S_\phi}\{GT^{p0}, P_n^{p0}\} + L_{perceptual,O_\phi}\{GT^{p0}, P_n^{p0}\}\right) \tag{6}$$

Where,

$N$ = Number of physician styles & $\lambda_{1n}^{p0} = \begin{cases} 1, & \text{patient} \in \text{Physician } n \\ 0, & \text{patient} \notin \text{Physician } n \end{cases}$

Addition of perceptual loss does not significantly improve model accuracy, but without the perceptual loss, the predicted contours from the 5 different decoders were visibly not very dissimilar from each other. This encoder-multidecoder approach can also be easily adapted to a scenario in which all the physician data is unavailable for training. If an existing segmentation model (could be mixed-style or a specific style model) has to be made style-aware for a new physician at the same institution or if a model trained with a single institution's data needs to be adapted to another institution's style, the architecture can be expanded easily with an additional decoder added.

The input to the model are the CT and corresponding bladder and rectum masks that were automatically generated from the CT using DL networks as outlined in Balagopal et al.[12] All the models were trained on a V100 GPU with 32 GB of memory, using TensorFlow version 2.1. The batch size was set to 1 due to memory limitations.

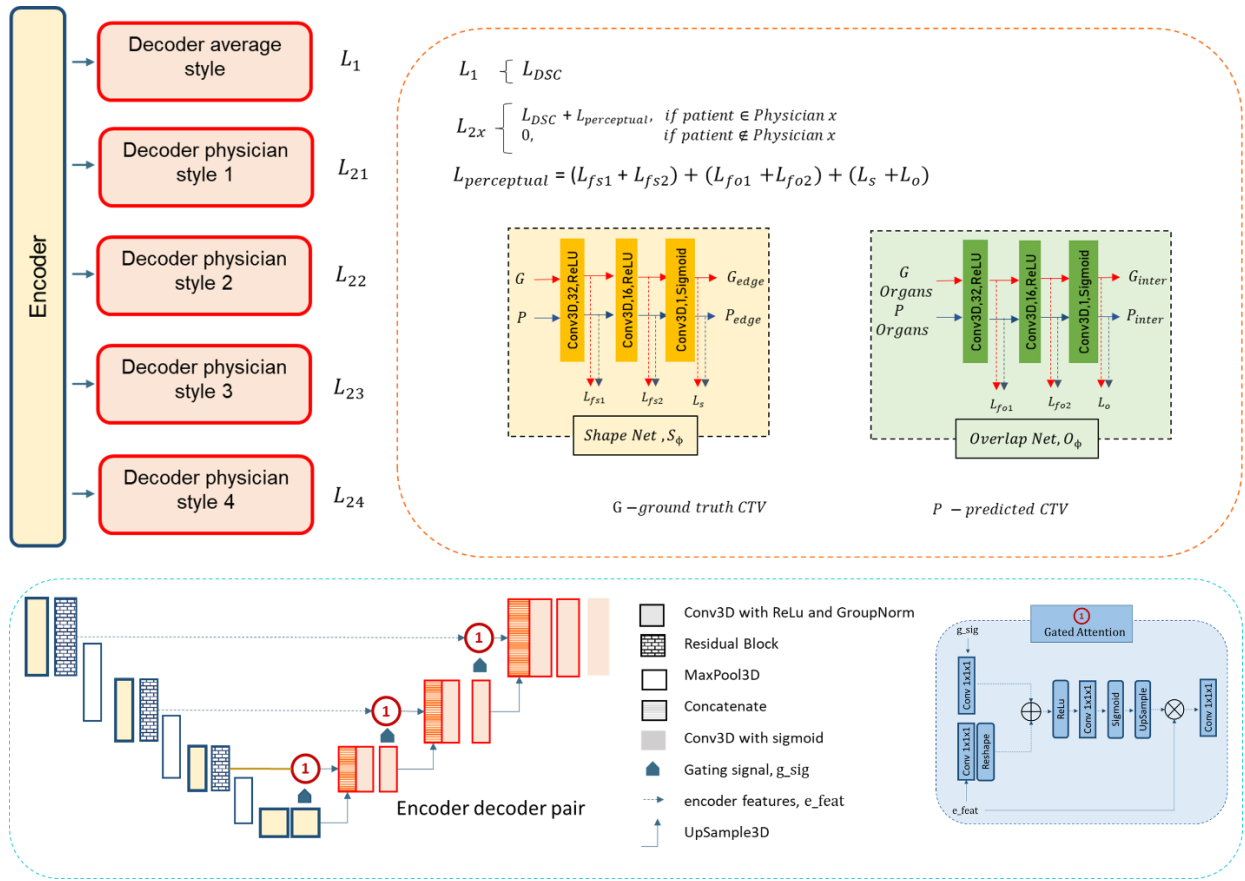

*Figure 4: Schematic of the Physician-style-aware network (PSA-Net) with feature attention and perceptual loss.*

## 2.4 OUTCOME ANALYSIS

Because physicians have stylistic contouring preferences that are distinguishable even when using similar contouring guidelines, it is important to estimate the effect of these style variations on patient outcomes. Table 1 shows baseline characteristics of the patients included in the outcome/toxicity analysis, obtained from retrospective chart review.

Biochemical failure after PORT was defined as PSA failure (prostate-specific antigen (PSA) ≥ nadir+2)[16]. Late toxicities following radiotherapy were defined as occurring at least 90 days after completion of radiotherapy and were graded with the Common Terminology Criteria for Adverse Events (CTCAE) version 5.0 scale. The endpoints overall survival (OS), biochemical recurrence-free survival (BCFS), distant metastasis-free survival (DMFS), GI toxicity-free survival (GIFS), and GU toxicity-free survival (GUFS) were defined as time until death due to any cause, biochemical recurrence, diagnosis of metastasis, diagnosis of grade ≥ 3 GI toxicity, and diagnosis of grade ≥ 3 GU toxicity, respectively (Table 2). Survival times were calculated from the date of radiotherapy completion until the date of the first occurrence of the endpoint or were censored at the date of last follow-up if the endpoint was not reached.

All statistical analyses were done in SPSS version 25 (IBM). Unless otherwise specified, a two-tailed $p<0.05$ was deemed statistically significant. The baseline characteristics of patients belonging to different physicians could play a significant role in patient outcome and hence the distributions have to be compared to evaluate if there are significant differences between physicians. Baseline characteristics of the patients were evaluated for differences between providers with the chi-squared or Fisher exact test for categorical variables and one-way ANOVA for continuous variables. We computed Kaplan-Meier survival curves for OS, BCFS, DMFS, GIFS, and GUFS. We evaluated differences in survival outcomes between providers with the log rank test. We then conducted a multivariate Cox regression analysis to evaluate the hazard ratios of the associated baseline characteristics for predicting BCFS.

| Variable | Number | Percentage (%) |
|---|---|---|
| **Physician** | | |
| 1 | 38 | 19.1% |
| 2 | 64 | 32.2% |
| 3 | 21 | 10.6% |
| 4 | 76 | 38.2% |
| **Primary Gleason Score** | | |
| 3 | 108 | 54.3% |
| 4 | 86 | 43.2% |
| 5 | 4 | 2.0% |
| NA | 1 | 0.5% |
| **Secondary Gleason Score** | | |
| 3 | 73 | 36.7% |
| 4 | 110 | 55.3% |
| 5 | 15 | 7.5% |
| NA | 1 | 0.5% |
| **Tertiary Gleason score** | | |
| 3 | 2 | 1.0% |
| 4 | 1 | 0.5% |
| 5 | 13 | 6.5% |
| NA | 183 | 92.0% |
| **Total Gleason score** | | |
| 6 | 8 | 4.0% |
| 7 | 165 | 82.9% |
| 8 | 8 | 4.0% |
| 9 | 18 | 9.0% |
| **Path T stage at surgery** | | |
| pT2a | 5 | 2.5% |
| pT2b | 3 | 1.5% |
| pT2c | 86 | 43.2% |
| pT3a | 68 | 34.2% |
| pT3b | 31 | 15.6% |
| pT4 | 4 | 2.0% |
| NA | 2 | 1.0% |
| **Nodal Stage** | | |
| 0 | 191 | 96.0% |
| 1 | 7 | 3.5% |
| NA | 1 | 0.5% |
| **Surgical Margins** | | |
| No | 68 | 34.2% |
| Yes | 129 | 64.8% |
| NA | 2 | 1.0% |
| **Salvage Radiation Total Dose (Gy)** | | |
| 66.4 | 1 | 0.5% |
| 66.6 | 39 | 19.6% |
| 68.4 | 8 | 4.0% |
| 70.2 | 145 | 72.9% |
| 70.4 | 1 | 0.5% |
| 72 | 5 | 2.5% |
| **Salvage Radiation Fractions** | | |
| 36 | 4 | 2.0% |
| 37 | 39 | 19.6% |
| 39 | 150 | 75.4% |
| 40 | 5 | 2.5% |
| 42 | 1 | 0.5% |
| **ADT received** | | |
| No | 110 | 55.3% |
| Yes | 87 | 43.7% |
| NA | 2 | 1.0% |

*Table 1: Baseline characteristics of the patient cohort*

| Outcome/Toxicity | Number | Percentage |
|---|---|---|
| **Biochemical Failure** | | |
| No | 162 | 81.4% |
| Yes | 37 | 18.6% |
| **Overall Survival Failure** | | |
| No | 198 | 99.5% |
| Yes | 1 | 0.5% |
| **Distant Metastasis** | | |
| No | 185 | 93.0% |
| Yes | 14 | 7.0% |
| **Grade >=3 GI toxicity** | | |
| No | 194 | 97.5% |
| Yes | 5 | 2.5% |
| **Grade >=3 GU toxicity** | | |
| No | 195 | 98.0% |
| Yes | 4 | 2.0% |

*Table 2: Number of events in the survival/toxicity outcomes for this study*

## 3. RESULTS

## 3.1 CLASSIFICATION NETWORK RESULTS

We trained the classifier on the data from 4 UTSW physicians. The model was able to classify all 4 styles, and accuracy on the test dataset was 87.2%. This result shows that the 4 physicians from UTSW in this study have 4 different DL distinguishable contouring styles.

## 3.2 OUTCOME ANALYSIS

**Baseline comparisons**

Table 3 summarizes results of the statistical tests used for evaluating differences between the baseline characteristics of patients across physicians. Distributions of nodal stage, total radiation dose, radiation fractions, and ADT were found to be significantly different between physicians.

| Variable | Tests | p-value |
|---|---|---|
| Primary Gleason score | Fisher-Freeman-Halton Exact Test | 0.25 |
| Secondary Gleason score | Fisher-Freeman-Halton Exact Test | 0.08 |
| Total Gleason score | Fisher-Freeman-Halton Exact Test | 0.59 |
| Nodal Stage | Fisher-Freeman-Halton Exact Test | 0.03 |
| Surgical Margins | Chi-Squared | 0.16 |
| ADT | Chi-Squared | .000 |
| Total Radiation Dose | One-way ANOVA | .013 |
| Radiation Fractions | One-way ANOVA | .004 |
| Pre-op PSA | One-way ANOVA | .414 |
| Post-op PSA | One-way ANOVA | .481 |
| Pre-salvage PSA | One-way ANOVA | .464 |
| Time from surgery to Radiation | One-way ANOVA | .984 |

*Table 3: Statistical comparison of baseline characteristics between providers*

## Kaplan-Meier Survival Analysis

The Kaplan-Meier curves for BCFS, DMS, GIFS and GUFS appear in Figure 5. The analysis could not be performed for OS, since there was only 1 patient that reached the endpoint of death during follow-up for this dataset. *p*-values for survival analysis were calculated using the logrank test. On logrank analysis, there were no statistically significant differences between providers for any clinical endpoint, including BCFS, DMFS, GIFS, or GUFS.

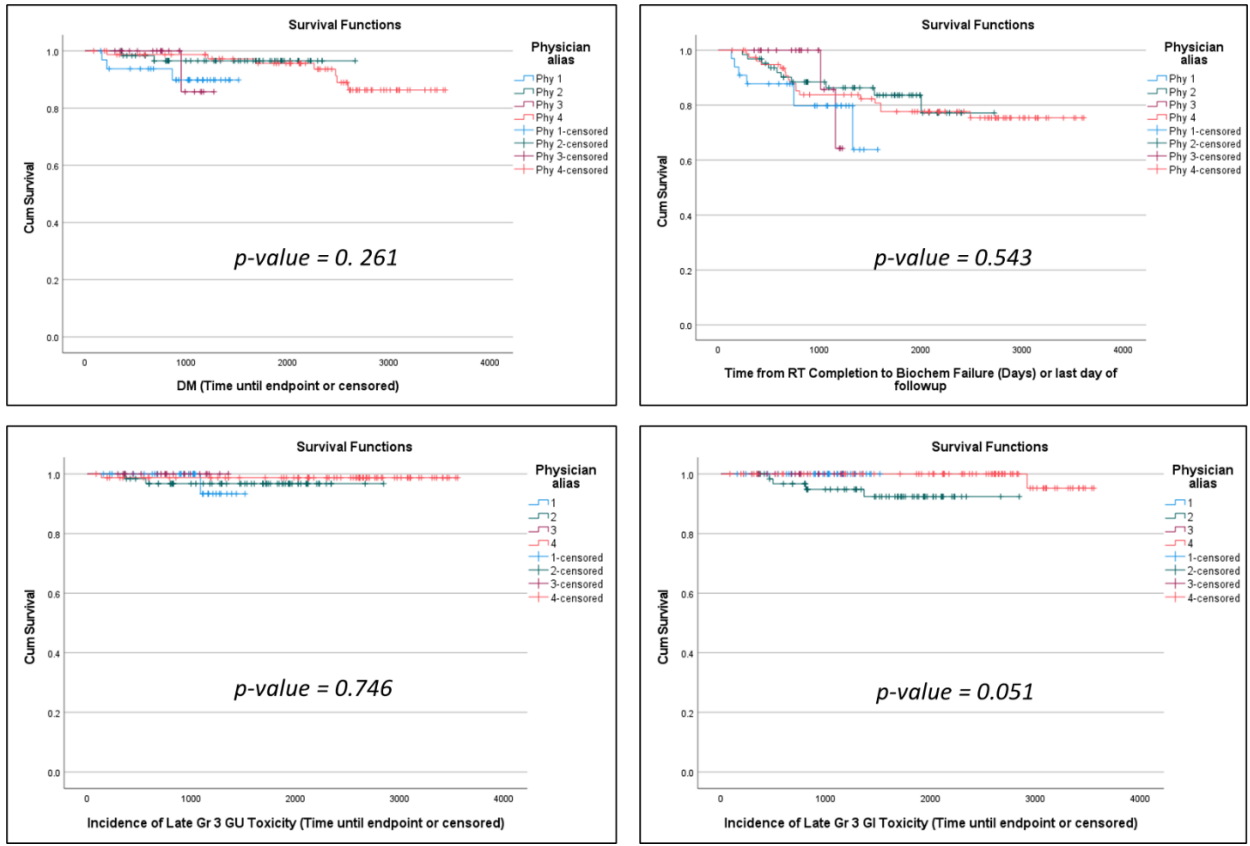

*Figure 5: Kaplan-Meir survival analysis plots for Biochemical Free Survival, Distant Metastasis Free Survival, Late Grade>=3 GI toxicity free survival and Late Grade >=3 GU toxicity free survival*

**Multivariate Cox Regression**

Similar to the log rank analysis, physicians were not observed to be a significant factor associated with BCFS on multivariate Cox regression analysis either. Surgical margins, nodal stage and Gleason scores were observed to be significant predictors of BCFS. Higher nodal stage and primary and tertiary Gleason scores were observed to be associated with higher risk of biochemical regression. Positive surgical margins were observed to have a negative co-efficient which signifies that negative surgical margins are significantly associated with increased biochemical regression. The results of the multivariate Cox regression analysis for BCFS is shown in Table 4.

| Variable | Coefficient (B) | SE | p-value | Hazard Ratio (HR) | 95% CI for HR (Lower) | 95% CI for HR (Upper) |
|---|---|---|---|---|---|---|
| Physician | - | - | .550 | - | - | - |
| **Nodal Stage** | 2.01 | .905 | .026 | 7.462 | 1.266 | 43.985 |
| **Surgical Margins** | -1.361 | .420 | .001 | 0.256 | 0.177 | 0.728 |
| Path T stage at surgery | - | - | .051 | - | - | - |
| ADT | - | - | .103 | - | - | - |
| Salvage Radiation Total Dose(Gy) | - | - | .629 | - | - | - |
| Salvage Radiation Fractions | - | - | .626 | - | - | - |
| **Primary Gleason** | .784 | .387 | .043 | 2.190 | 1.026 | 4.672 |
| Secondary Gleason | - | - | .933 | - | - | - |
| **Tertiary Gleason** | 1.115 | .852 | .008 | 3.051 | .575 | 16.195 |
| Post-op PSA | - | - | .347 | - | - | - |
| Pre-salvage PSA | - | - | .264 | - | - | - |

*Table 4: Cox regression analysis result for BCFS*

## 3.3 STYLE ADAPTATION

**Quantitative Analysis**

For evaluating style adaptation, Dice similarity coefficients were compared; see Figure 6a. We compared four models: A mixed style segmentation model (Model A1) trained on all the data; Physician sub-model (Model A2) trained on only a single physician data; Physician TL-adapted model (Model A3), the mixed style segmentation model adapted to each physician using transfer learning; and the style-aware segmentation model (Model A4). DSC mean and standard deviation for each of the physicians is tabulated in Figure 6b for each of the four models compared. Mixed-style segmentation has an average test accuracy of 87.8%, 82.8%, 83.3% and 80.0% for classifying contours correctly for Physicians 1, 2, 3 and 4, respectively. Style-aware segmentation improves the accuracy by an average of 3%. Physician sub-models performed worse compared to the mixed-style model and the style-aware model. The physician TL-adapted model performed better than the mixed-style model and the physician sub-models, but worse than the style-aware segmentation model. A paired *t* test for means was used to evaluate significance of style-aware segmentation model; the *p*-values are reported in Figure 6b. Comparison of distance metrics, Hausdorff distance 95% percentile (HD95), average surface distance (ASD), and average symmetric surface distance (ASSD) along with True Positive Ratio (TPR) and True Negative Ratio (TNR) are reported in Figure 6c. When comparing distance metrics, the style-aware segmentation model performs much better than the other three models, with a higher mean values and lower standard deviation.

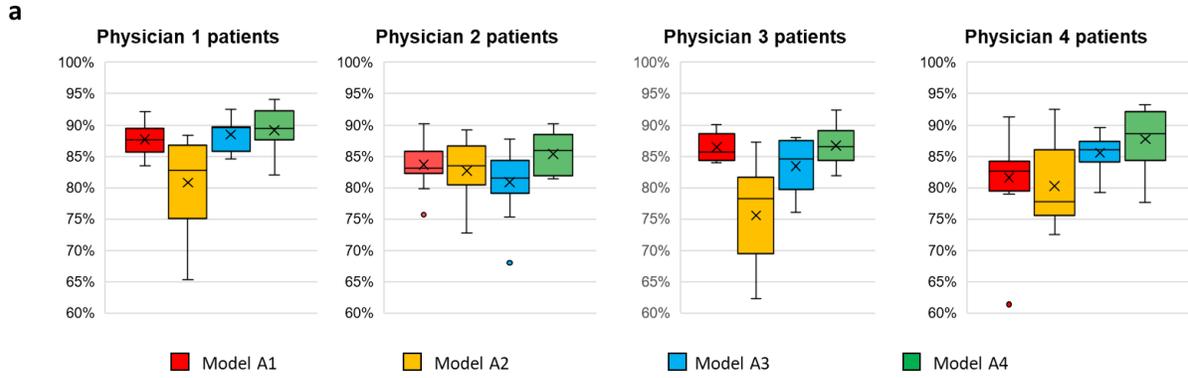

*Figure 6: Physician style-aware segmentation result comparison for mixed style segmentation model (Model A1),which is trained on all the data, Physician sub model (Model A2),which is trained on only a single corresponding physician data, Physician TL-adapted model (Model A3), which is the mixed style segmentation model adapted to each physician using transfer learning, and the style-aware segmentation model (Model A4) a) Box-plots comparing DSC values of each model for patients contoured by each physician. b) Mean DSC values along with standard deviation for the plots in a along with p-value for significance comparison between mixed-style segmentation model and style-aware segmentation model. c) Distance metrics, True positive rate, and True negative rate for each model on all test data.*

## Visual analysis

With the use of perceptual loss, the physician style-aware contours were visibly different and looked closer to physician contours. A few examples that show these visual differences are picturized in Figure 7. Most differences between the physician style segmentations and mixed-style segmentations were stylistic. Both produce accurate, acceptable contours, with the style-aware segmentations showing an improved similarity with physician contours.

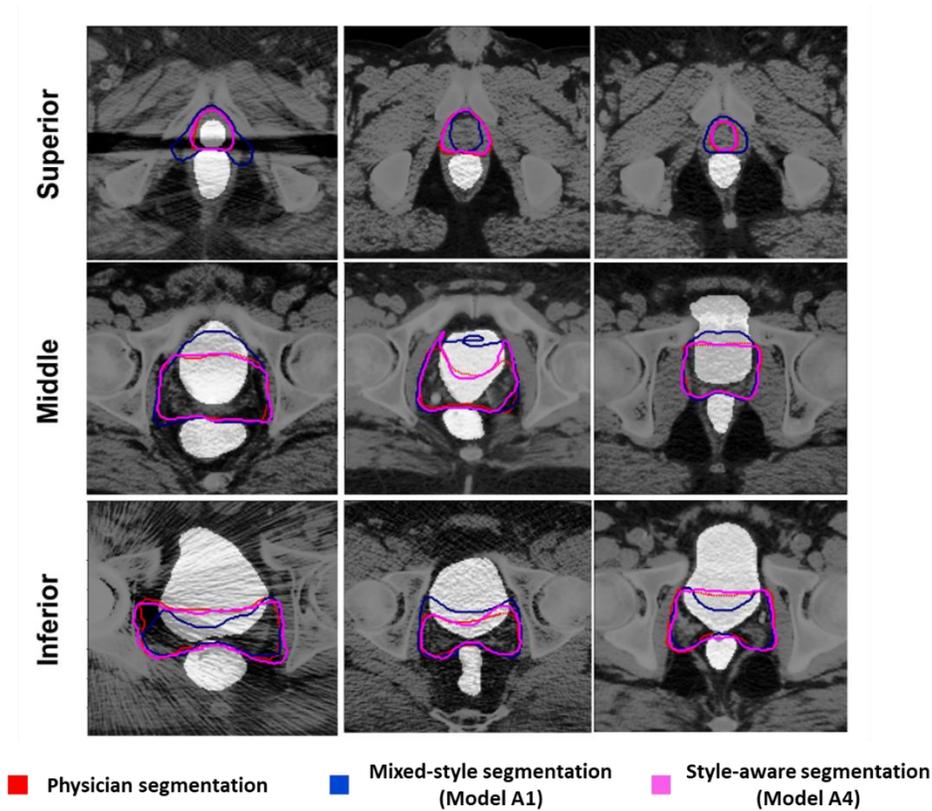

*Figure 7: Visual depiction of style-aware segmentation on different 2D slices extracted from different patients at inferior, middle and superior positions. Majority of the improvements were on the middle slices where bladder inclusion and CTV shapes were visibly different for the physicians.*

**Mayo dataset – inter institutional style variation**

We evaluated the performance of the post-operative prostate CTV segmentation model trained on the UTSW dataset (Model A) by applying it to the Mayo dataset. Because DL generated Bladder and Rectum are required inputs for the CTV model, we also evaluated the accuracy of OAR segmentation models on the entire dataset. The accuracy in terms of DSC for Bladder and Rectum were 94.4 (± 3.5) % and 87.5 (± 6.2) % respectively. The accuracy for CTV meanwhile, was only 70.5 (± 6.4) % owing to large variations in CTV contouring between the two institutions.

To evaluate institutional-style adaptation and physician-style adaptation, the data was split into training (53 patients) and testing (30 patients contoured by a single physician). Three models were trained. Model B was trained from scratch with only Mayo data used for training. Model C, an institutional style-aware model was trained using transfer learning from Model A. The encoder was frozen (weights not updated during training) and a new decoder, initialized with weights from the UTSW model was retrained with the new data. Model D was trained with 2 additional decoders added to the frozen Model A encoder. One decoder branch learned mixed-style features and the second decoder learned physician 1 style-aware segmentation as outlined in 2.3. Physician-style aware segmentations could not be generated for the other 5 physicians due to the modest numbers of patients available for each. A comparison for the four models is plotted in Figure 8a. Model B, trained only on the Mayo dataset, performed better than model A. The institutional style-adapted model (model C), performed better than the model trained from scratch on the Mayo dataset. Model D (physician1 style-aware model) performed the best.

Comparisons of distance metrics, Hausdorff distance, average surface distance, average symmetric surface distance, TPR, and TNR are reported in Figure 8b. When comparing distance metrics, the style-aware segmentation model performs much better than the other three, with higher mean and lower standard deviation. Figure 8c shows a visual comparison between the model A and Model D for a few inferior, middle, and superior slices for two test patients. While Model A contours are considered clinically acceptable at UTSW, they are unacceptable at Mayo. Similarly, Model D contours are more acceptable (better agreement with ground truth) at Mayo clinic whose data was used for training Model D.

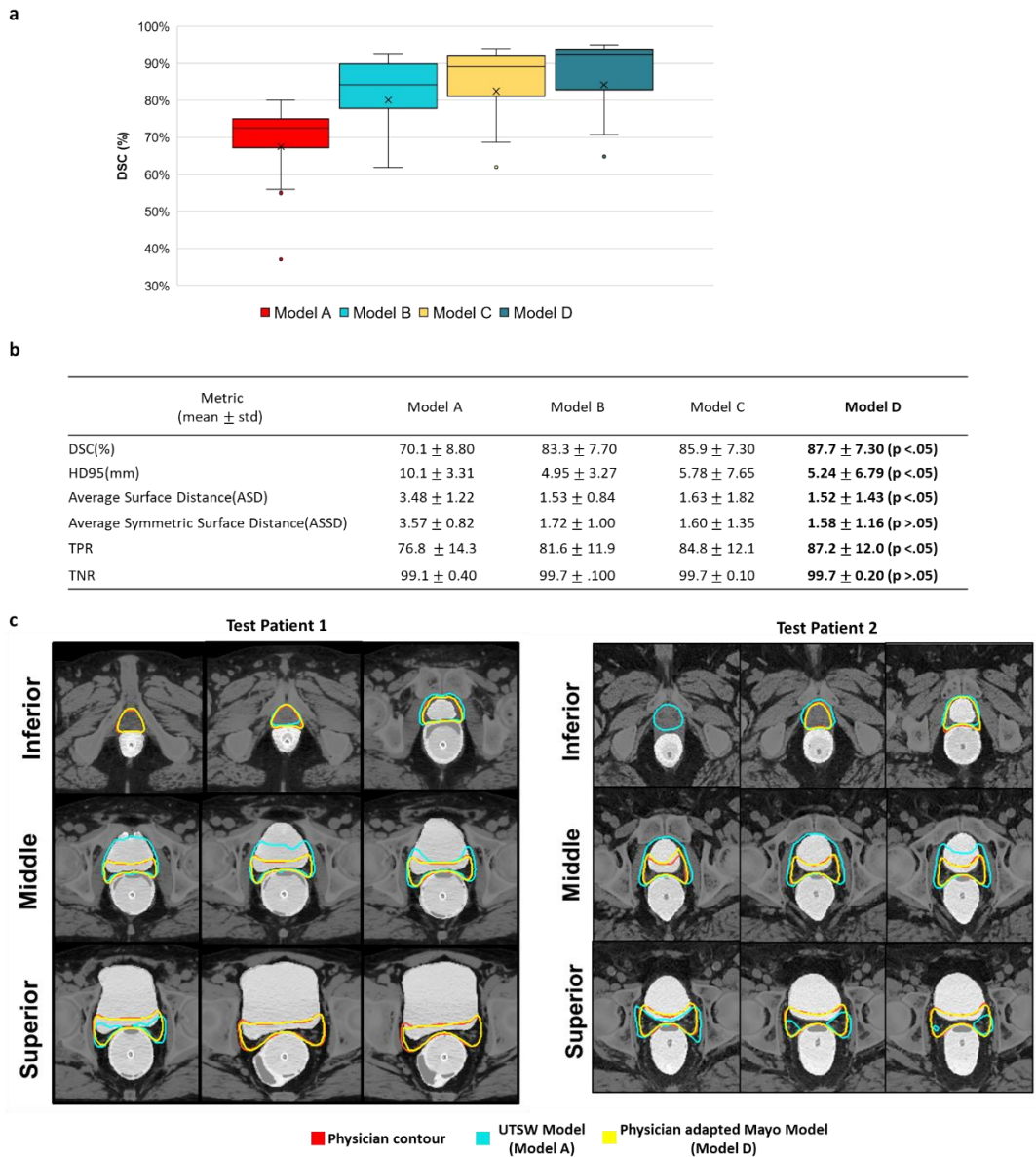

a)

b)

| Metric (mean ± std) | Model A | Model B | Model C | Model D |
|---|---|---|---|---|
| DSC(%) | 70.1 ± 8.80 | 83.3 ± 7.70 | 85.9 ± 7.30 | **87.7 ± 7.30 (p <.05)** |
| HD95(mm) | 10.1 ± 3.31 | 4.95 ± 3.27 | 5.78 ± 7.65 | **5.24 ± 6.79 (p <.05)** |
| Average Surface Distance(ASD) | 3.48 ± 1.22 | 1.53 ± 0.84 | 1.63 ± 1.82 | **1.52 ± 1.43 (p <.05)** |
| Average Symmetric Surface Distance(ASSD) | 3.57 ± 0.82 | 1.72 ± 1.00 | 1.60 ± 1.35 | **1.58 ± 1.16 (p >.05)** |
| TPR | 76.8 ± 14.3 | 81.6 ± 11.9 | 84.8 ± 12.1 | **87.2 ± 12.0 (p <.05)** |
| TNR | 99.1 ± 0.40 | 99.7 ± .100 | 99.7 ± 0.10 | **99.7 ± 0.20 (p >.05)** |

c)

*Figure 8: Performance on Mayo dataset. a) Box plot comparing DSC result for Model A (Trained on UTSW dataset), Model B (Trained on Mayo dataset), Model C (Model adapted from UTSW model using Mayo dataset) and Model D (Physician 1 style-aware segmentation model). b) Distance metrics, True positive rate and True negative rate for each*

*model on test data c) visual comparison between the ground truth physician contours, Model A contours and Model D contours for a few inferior, middle and superior slices for two test patients.*

## 4. DISCUSSION

Inter-observer variability is a known problem in medical image segmentation. Variation is more prominent in the case of postoperative CTV segmentation due to the absence of a visual target on images with clear boundaries. Although this problem has been acknowledged, no attempts have been made to deal with it in a clinically meaningful way. In this work, we propose a novel physician style-aware approach capable of mimicking different styles of CTV segmentation if they exist. Such a technique has the potential to provide better acceptance of automatic methods in clinical practice, as it produces a set of predicted contours from which a physician can choose the preferred one. We provide proof of principle for our approach for postoperative prostate CTV segmentation data with known inter-observer variations on CT scans. We show that a DL model should be able to mimic the physician styles by successfully training a classifier that can identify the treating physician from the CTV contour. The classifier had an accuracy of 87% on the test dataset.

Given concern that stylistic differences in contouring styles could affect the long-term biochemical control or toxicity, we conducted an outcome/toxicity study for patients who underwent radiation to the prostate bed alone to evaluate if it would be clinically feasible to adapt segmentation algorithms to stylistic differences. Log rank tests showed no significant differences between providers for BCFS or any other outcome assessed, including DMFS, GIFS, or GUFS. On subsequent multivariate Cox regression, there was still no significant association between providers on BCFS, suggesting that the stylistic differences between providers did not have a significant impact on biochemical control. Positive nodal stage, higher primary as well as tertiary Gleason scores and negative surgical margins were observed to be associated with higher risk of biochemical failure. These results are in accordance with some previous publications[17,18]. Higher pre-SRT PSA, higher Gleason scores, negative surgical margins, lack of ADT, and lower SRT dose have been observed to be significantly associated with increased biochemical failure. It should be noted that for our study, the data is limited in its retrospective nature and that some baseline characteristics differed between groups which are known to affect biochemical control, including nodal stage, ADT, and the radiation prescription dose. Although GI toxicity was worse with physician 2, this was based on a very low incidence of reported grade 3+ GI events (n=5), with 4 being for physician 2, limiting meaningful comparisons between groups for toxicity assessment. Still, this preliminary data suggests that as long as physicians work within the accepted guidelines[1-4], they have the freedom to decide the CTV contour that is apt for their patients' treatment. Therefore, from a clinical standpoint, the adaptation of automatically segmented contours to the specific style of the physician via a DL approach appears feasible. However, we caution the interpretation of this data, given that the stylistic differences evaluated were among experienced radiation oncologists specializing in GU malignancies and may not be applicable to the inexperienced physician.

In the current scenario, the goal of an AI-segmentation algorithm is not to replace the role of physicians in contouring but to assist them in performing it more efficiently. AI algorithms could save contouring time and free up physicians to perform other tasks. Because the goal is to reduce contouring time, any approach that could make the AI contours more acceptable by physicians would greatly benefit the clinic. Given that stylistic variations exist, and a ground truth CTV cannot be established in a post-operative setting, physicians have the upper hand in deciding what to include in the CTV as long as it is within the guidelines. Performing physician-style aware segmentation given the scenario could reduce the amount of contour editing time by physicians and hasten patient treatment. With multiple physician-style automated contour predictions available for their patients, physicians would also be able to compare and contrast different CTV possibilities, understand possible regions of variation, and reduce variability in CTV contouring. When used in the clinic, the attending physician would have his default contour available. Along with their contours, the physician could refer to how other physicians would contour for this patient and make an informed judgement.

The proposed encoder-multidecoder approach, PSA-Net, is novel and makes use of a new perceptual loss function that produces visually different CTVs for each physician that mimics their style accurately. We also compared the proposed model with 1) a general model that is not style-aware, 2) models trained individually on each physician data and 3) a model trained on all data and transfer-learned to each physician's data. Our algorithm shows superior results on both UTSW and Mayo datasets with a DSC improvement greater than 3%. Visual inspection of the CTV contours further validated the better conformity of style adapted contours compared to mixed-style contours.

We also evaluated our approach on an external dataset acquired at Mayo clinic. Even though both institutions use the RTOG CTV contouring guidelines, we observed large and consistent variations between the institutions. This shows the necessity of style-adaptation not only to improve segmentation accuracy within an institution but also across institutions given the large differences in contouring styles.

It has to be considered that all the physicians included in this study are experienced in the field. When it comes to CTV segmentation, two types of physician variations can be considered. One that is undesirable and caused by inexperienced physicians. Training a model to mimic an inexperienced physician's style would not be advantageous. This study focuses on a second variant, which is the variability that exists within a group of experienced physicians. Given the absence of a ground truth, clinical judgement of every physician has to be given equal weight unless a proof of the contrary is available. With the available data, we have conducted a preliminary study that suggested that the stylistic differences between providers did not have a significant impact on biochemical control. This study is by no means comprehensive, as the number of post-operative patients available was modest. The statistical analysis done showed that the style differences led to no differences in outcomes; thus, for the moment, there is no compelling reason to force physicians to be completely uniform in contouring CTVs for the postoperative prostate cancer patients.

The main assumption of this approach is that there exist consistent, learnable segmentation style differences between physicians. However, in the event that observers are inconsistent where the variations are random (physician does not have a consistent learnable style), the applicability of our approach is limited and needs further study. Another assumption is that the labels of physicians who have contoured each patient are required for training. This information is not difficult to come by, since treating physician information is commonly archived in the metadata of the CT simulation DICOM files. In situations where such information is unavailable, this approach cannot be used. Also, the existence of clean contouring style differences in not a necessity at all institutions. In some places, institution-specific contouring guidelines exists that mandate uniform CTVs across all physicians, but such institutions are rare. In those cases, a general CTV segmentation model would suffice. A classification model can greatly help understand whether there exist consistent learnable segmentation style differences between physicians.

## 5. CONCLUSIONS

Inter-observer variation is a consistent limitation in medical image segmentation. Current CNN networks are unable to learn style-aware segmentation given data from multiple physicians, producing mixed-style segmentation. In this study, the performance of the classification network demonstrated the feasibility of learning physician styles and incorporating them in a segmentation model. Additionally, adapting contours to the specific styles of physicians contouring within the bounds of accepted guidelines via a DL approach appears to be clinically feasible, given the lack of an association between contouring style and BCFS in this study. The proposed PSA-Net is capable of producing contours specific to the treating physician, improving segmentation accuracy, and thus, avoiding the need to train multiple models to achieve different style segmentations. PSA-Net was further validated on data from a separate institution, demonstrating the ease of adaptation to new physicians' styles when calibrated with their cases. PSA-Net is a promising new framework that after further development and validation on larger datasets, may improve the clinic workflow by decreasing time spent on contouring and editing and further provide practitioners the ability to view how a target volume may be contoured by a prominent physician via a mouse click for educational purposes.

**Acknowledgement**

We would like to thank Dr. Jonathan Feinberg for editing the manuscript and Varian Medical Systems, Inc. for providing funding support.

**Funding**

Funding support was provided by the Varian Medical systems' research grant.

**Declaration of Competing Interests**

The authors declare that there are no competing interests.

## REFERENCES

1. Poortmans P, Bossi A, Vandeputte K, Bosset M, Miralbell R, Maingon P, et al. Guidelines for target volume definition in post-operative radiotherapy for prostate cancer, on behalf of the EORTC Radiation Oncology Group. Radiother Oncol (2007) 84(2):121–7.10.1016/j.radonc.2007.07.017
2. Sidhom MA, Kneebone AB, Lehman M, Wiltshire KL, Millar JL, Mukherjee RK, et al. Post-prostatectomy radiation therapy: consensus guidelines of the Australian and New Zealand Radiation Oncology Genito-Urinary Group. Radiother Oncol (2008) 88(1):10–9.10.1016/j.radonc.2008.05.006
3. Wiltshire KL, Brock KK, Haider MA, Zwahlen D, Kong V, Chan E, et al. Anatomic boundaries of the clinical target volume (prostate bed) after radical prostatectomy. Int J Radiat Oncol Biol Phys (2007) 69(4):1090–9.10.1016/j.ijrobp.2007.04.068
4. Michalski JM, Lawton C, El Naqa I, Ritter M, O'Meara E, Seider MJ, et al. Development of RTOG consensus guidelines for the definition of the clinical target volume for postoperative conformal radiation therapy for prostate cancer. Int J Radiat Oncol Biol Phys (2010) 76(2):361–8.10.1016/j.ijrobp.2009.02.006

5. Sophie Robin, Marjory Jolicoeur, Samuel Palumbo, Thomas Zilli, Gilles Crehange, Olivier De Hertogh, Talar Derashodian, Paul Sargos, Carl Salembier, Stéphane Supiot, Corina Udrescu, Olivier Chapet, Prostate Bed Delineation Guidelines for Postoperative Radiation Therapy: On Behalf Of The Francophone Group of Urological Radiation Therapy, International Journal of Radiation Oncology*Biology*Physics, 2020, ISSN 0360-3016, https://doi.org/10.1016/j.ijrobp.2020.11.010. (https://www.sciencedirect.com/science/article/pii/S0360301620344989)
6. Mitchell DM, Perry L, Smith S, Elliott T, Wylie JP, Cowan RA, Livsey JE, Logue JP. Assessing the effect of a contouring protocol on postprostatectomy radiotherapy clinical target volumes and inter-physician variation. Int J Radiat Oncol Biol Phys. 2009;75(4):990–993. doi: 10.1016/j.ijrobp.2008.12.042.
7. Lawton CA, Michalski J, El-Naqa I, Kuban D, Lee WR, Rosenthal SA, Zietman A, Sandler H, Shipley W, Ritter M. et al. Variation in the definition of clinical target volumes for pelvic nodal conformal radiation therapy for prostate cancer. Int J Radiat Oncol Biol Phys. 2009;74(2):377–382. doi: 10.1016/j.ijrobp.2008.08.003.
8. Lawton CA, Michalski J, El-Naqa I, Buyyounouski MK, Lee WR, Menard C, O'Meara E, Rosenthal SA, Ritter M, Seider M. RTOG GU Radiation oncology specialists reach consensus on pelvic lymph node volumes for high-risk prostate cancer. Int J Radiat Oncol Biol Phys. 2009;74(2):383–387. doi: 10.1016/j.ijrobp.2008.08.002.
9. Livsey JE, Wylie JP, Swindell R, Khoo VS, Cowan RA, Logue JP. Do differences in target volume definition in prostate cancer lead to clinically relevant differences in normal tissue toxicity? Int J Radiat Oncol Biol Phys. 2004;60(4):1076–1081. doi: 10.1016/j.ijrobp.2004.05.005.
10. Symon Z, Tsvang L, Pfeffer MR, Corn B, Wygoda M, Ben-Yoseph R. Prostatic fossa boost volume definition: physician bias and the risk of planned geographical miss. In: Proceedings 90th annual RSNA meeting, Chicago; 2004. Abstract SSC19-09.
11. Lee E, Park W, Ahn SH, et al. Interobserver variation in target volume for salvage radiotherapy in recurrent prostate cancer patients after radical prostatectomy using CT versus combined CT and MRI: a multicenter study (KROG 13-11). Radiat Oncol J. 2017;36(1):11–16. doi:10.3857/roj.2017.00080
12. Balagopal A, Nguyen D, Morgan H et al. A deep learning-based framework for segmenting invisible clinical target volumes with estimated uncertainties for post-operative prostate cancer radiotherapy**.** arXiv preprint arXiv: 2004.13294. 2020.
13. J. Johnson, A. Alahi, and L. Fei-Fei. Perceptual losses for real-time style transfer and super-resolution. In European conference on computer vision, pages 694–711. Springer, 2016.
14. K. Simonyan and A. Zisserman. Very deep convolutional networks for large-scale image recognition. In International Conference on Learning Representations, 2015.
15. Yifu Chen, Arnaud Dapogny, Matthieu Cord. SEMEDA: Enhancing segmentation precision with semantic edge aware loss, Pattern Recognition, Volume 108, 2020,107557, ISSN 0031-3203, https://doi.org/10.1016/j.patcog.2020.107557.
16. Roach M 3rd, Hanks G, Thames H Jr, Schellhammer P, Shipley WU, Sokol GH, Sandler H. Defining biochemical failure following radiotherapy with or without hormonal therapy in men with clinically localized prostate cancer: recommendations of the RTOG-ASTRO Phoenix Consensus Conference. Int J Radiat Oncol Biol Phys. 2006 Jul 15;65(4):965-74. doi: 10.1016/j.ijrobp.2006.04.029. PMID: 16798415
17. Postoperative radiotherapy after radical prostatectomy for high-risk prostate cancer: long-term results of a randomised controlled trial (EORTC trial 22911), The Lancet,Volume 380, Issue 9858,

2012, Pages 2018-2027,ISSN 0140-6736,https://doi.org/10.1016/S0140-6736(12)61253-7.(https://www.sciencedirect.com/science/article/pii/S0140673612612537).
18. Rahul D. Tendulkar, Shree Agrawal, Tianming Gaoet al., Contemporary Update of a Multi-Institutional Predictive Nomogram for Salvage Radiotherapy After Radical Prostatectomy. Journal of Clinical Oncology 2016 34:30, 3648-3654.